\documentclass[twoside,11pt]{article}

\usepackage{jmlr2e}

\usepackage{hyperref}
\usepackage{url}

\hypersetup{
  colorlinks=true,
  urlcolor={blue},
  citecolor={.},
}

\usepackage{multirow}
\usepackage{wrapfig}
\usepackage[export]{adjustbox}
\usepackage{pie_chart}

\usepackage[list=true]{subcaption}
\usepackage{amsmath}
\usepackage{mathtools}
\usepackage[pdf]{graphviz}

\usepackage{pgfplots}
\usepackage{pgfplotstable}
\pgfplotsset{width=10cm,compat=1.9}
\usepackage[capitalise]{cleveref}

\def\floor#1{\lfloor #1 \rfloor}

\renewcommand{\vec}[1]{\mathbf{#1}}

\newcommand{\mult}{\;.\;}

\newcommand{\norm}[1]{\left\lVert#1\right\rVert}

\newcommand{\Loss}{\mathcal{L}}

\newcommand{\Lagent}{\Loss_{imit}}
\newcommand{\Lenv}{\Loss_{env}}
\newcommand{\wagent}{{w}_{imit}}
\newcommand{\wenv}{{w}_{env}}

\newcommand{\M}[1]{\mathcal{M}_#1}

\newcommand{\site}{https://sites.google.com/view/waymo-learn-to-drive/}

\ShortHeadings{ChauffeurNet: Learning to drive by imitating the best and synthesizing the worst}{Bansal, Krizhevsky \& Ogale}
\firstpageno{1}

\begin{document}

\title{ChauffeurNet: Learning to Drive\\ by Imitating the Best and Synthesizing the Worst}

\author{\name Mayank Bansal \email mayban@waymo.com \\
       \addr Waymo Research\\
       Mountain View, CA 94043, USA\\
       \AND
       \name Alex Krizhevsky\thanks{Work done while at Google Brain \& Waymo.} \email akrizhevsky@gmail.com \\
       \AND
       \name Abhijit Ogale \email ogale@waymo.com \\
       \addr Waymo Research\\
       Mountain View, CA 94043, USA
       }

\editor{}

\maketitle

\begin{abstract}
Our goal is to train a policy for autonomous driving via imitation learning that is robust enough to drive a real vehicle.
We find that standard behavior cloning is insufficient for handling complex driving scenarios, even when we leverage a
perception system for preprocessing the input and a controller for executing the output on the car: 30 million examples are still not enough.
We propose exposing the learner to synthesized data in the form of perturbations to the expert's driving, which creates interesting situations
such as collisions and/or going off the road. Rather than purely imitating all data, we augment the imitation loss with additional losses
that penalize undesirable events and encourage progress -- the perturbations then provide an important signal for these losses and lead
to robustness of the learned model. We show that the {\em ChauffeurNet} model can handle complex situations in simulation, and present ablation
experiments that emphasize the importance of each of our proposed changes and show that the model is responding to the appropriate causal factors.
Finally, we demonstrate the model \href{\site}{driving} a car in the real world.

\end{abstract}

\begin{keywords}
  Deep Learning, Mid-to-mid Driving, Learning to Drive, Trajectory Prediction.
\end{keywords}

\section{Introduction}

In order to drive a car, a driver needs to see and understand the
various objects in the environment, predict their possible future
behaviors and interactions, and then plan how to control the car in
order to safely move closer to their desired destination while obeying
the rules of the road. This is a difficult robotics challenge that humans solve
well, making imitation learning a promising approach. Our work is about getting imitation learning to the level where it has a shot at driving a real vehicle; although the same insights may apply to other domains,
these domains might have different constraints and opportunities, so we do not want to claim contributions there.

We built our system based on leveraging the training data (30 million real-world expert driving examples, corresponding to about 60 days of continual driving) as effectively as possible. There is a lot of excitement for end-to-end learning approaches to driving
which typically focus on learning to directly predict raw control outputs such as steering or
braking after consuming raw sensor input such as camera or lidar
data. But to reduce sample complexity, we opt for mid-level input and output representations that take advantage of perception and control components. We use a perception system that processes raw sensor information and produces our input: a top-down representation
of the environment and intended route, where objects such as vehicles are
drawn as oriented 2D boxes along with a rendering of the road information and traffic
light states. We present this mid-level
input to a recurrent neural network (RNN), named {\em ChauffeurNet}, which then outputs a driving trajectory
that is consumed by a controller which translates it to steering
and acceleration. The further advantage of these mid-level representations
is that the net can be trained on real or simulated data, and can be easily tested and
validated in closed-loop simulations before running on a real car.

Our first finding is that even with 30 million examples, and even with mid-level input and output representations that remove the burden of perception and control, pure imitation learning is not sufficient. As an example, we found that this model would get stuck
or collide with another vehicle parked on the side of a narrow street, when a nudging
and passing behavior was viable. The key challenge is that we need to run the system closed-loop, where errors accumulate and induce a shift from the training distribution (\cite{ross2011reduction}). Scientifically, this result is valuable evidence
about the limitations of pure imitation in the driving domain, especially in light of recent promising results for high-capacity models (\cite{laskey2017comparing}). But practically, we needed ways to address this
challenge without exposing demonstrators to new states actively (\cite{ross2011reduction,laskey2017dart}) or performing reinforcement learning (\cite{kuefler2017imitating}).

We find that this challenge is surmountable if we augment the imitation loss with losses that discourage bad behavior and encourage progress, and, importantly,
augment our data with {\em synthesized} perturbations in the driving trajectory. These expose the model to non-expert behavior such as collisions and off-road driving,
and inform the added losses, teaching the model to avoid these behaviors. Note that the opportunity to synthesize this data comes from the mid-level input-output representations,
as perturbations would be difficult to generate with either raw sensor input or direct controller outputs.

We evaluate our system, as well as the relative importance of both loss augmentation and data augmentation, first in simulation. We then show how our final model
successfully drives a car in the real world and is able to negotiate situations involving
other agents, turns, stop signs, and traffic lights.
Finally, it is important to note that there are highly
interactive situations such as merging which may require a significant degree of
exploration within a reinforcement learning (RL) framework. This will demand simulating other (human) traffic participants, a rich area of ongoing research. Our contribution can be viewed as pushing the
boundaries of what you can do with purely offline data and no RL.

\section{Related Work \label{sec:related}}
Decades-old work on ALVINN (\cite{pomerleau1989alvinn}) showed
how a shallow neural network could follow the road by directly consuming
camera and laser range data. Learning to drive in an end-to-end manner has seen
a resurgence in recent years. Recent work by \cite{chen2015deepdriving}
demonstrated a convolutional net to estimate affordances such as distance
to the preceding car that could be used to program a controller to
control the car on the highway. Researchers at NVIDIA (\cite{bojarski2016end,bojarski2017explaining})
showed how to train an end-to-end deep convolutional neural network
that steers a car by consuming camera input. \cite{xu2017end} trained
a neural network for predicting discrete or continuous actions also
based on camera inputs. \cite{codevilla2018end} also train a network
using camera inputs and conditioned on high-level commands to output
steering and acceleration. \cite{kuefler2017imitating} use Generative
Adversarial Imitation Learning (GAIL) with simple affordance-style
features as inputs to overcome cascading errors typically present
in behavior cloned policies so that they are more robust to perturbations.
Recent work from \cite{hecker2018learning} learns a driving model
using 360-degree camera inputs and desired route planner to predict
steering and speed. The CARLA simulator (\cite{dosovitskiy2017carla})
has enabled recent work such as \cite{sauer2018conditional}, which
estimates several affordances from sensor inputs to drive a car in
a simulated urban environment. Using mid-level representations in a spirit
similar to our own, \cite{muller2018driving} train a system in simulation
using CARLA by training a driving policy from a scene segmentation
network to output high-level control, thereby enabling transfer learning
to the real world using a different segmentation network trained on
real data. \cite{pan2017virtual} also describes achieving transfer
of an agent trained in simulation to the real world using a learned
intermediate scene labeling representation. Reinforcement learning
may also be used in a simulator to train drivers on difficult interactive
tasks such as merging which require a lot of exploration, as shown
in \cite{shalev2016safe}. A convolutional network operating on a space-time volume of bird's eye-view
representations is also employed by \cite{fandf2018, djuric2018motion, lee2017convolution} for tasks like 3D detection, tracking
and motion forecasting. Finally, there exists a large volume of work on vehicle motion
planning outside the machine learning context and \cite{paden2016survey} present a notable survey.

\section{Model Architecture\label{sec:arch}}
\def\w{0.22}
\begin{figure*}[!t]
\centering
\subcaptionbox{Roadmap}{\includegraphics[width=\w\textwidth]{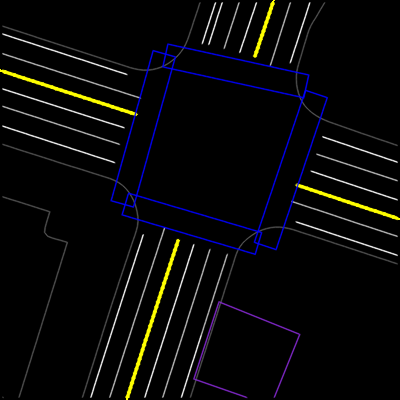}}
\hfill
\subcaptionbox{Traffic Lights}{
  \begin{tikzpicture}
        \node[anchor=south west,inner sep=0] (image) at (0,0) {\includegraphics[width=\w\textwidth,cframe=white]{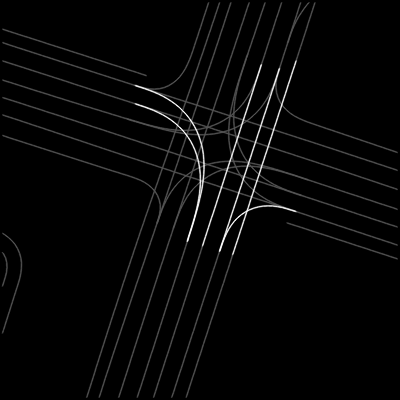}};
        \begin{scope}[x={(image.south east)},y={(image.north west)}]
            \node[anchor=south west,inner sep=0] (image) at (0.025,-0.025) {\includegraphics[width=\w\textwidth,cframe=white]{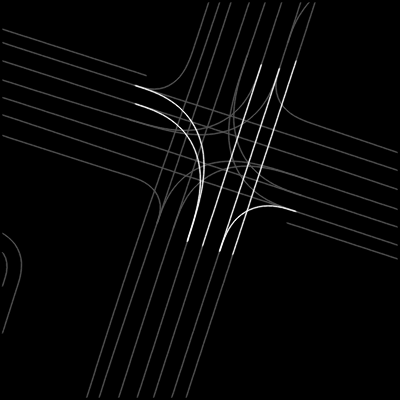}};
            \node[anchor=south west,inner sep=0] (image) at (0.05,-0.05) {\includegraphics[width=\w\textwidth,cframe=white]{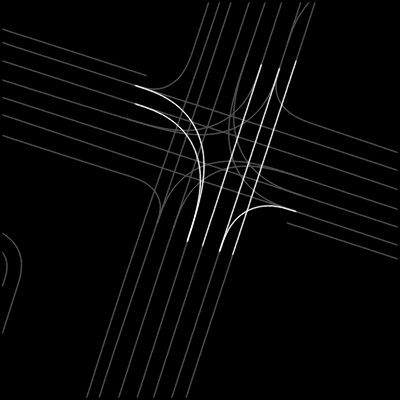}};
            \node[anchor=south west,inner sep=0] (image) at (0.075,-0.075) {\includegraphics[width=\w\textwidth,cframe=white]{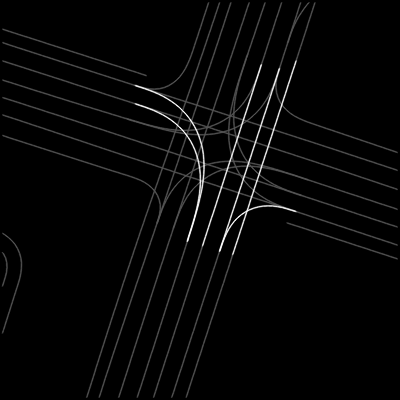}};
            \node[anchor=south west,inner sep=0] (image) at (0.1,-0.1) {\includegraphics[width=\w\textwidth,cframe=white]{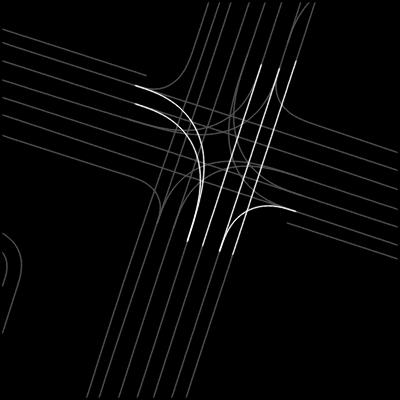}};
        \end{scope}
  \end{tikzpicture}
}
\hfill
\subcaptionbox{Speed Limit}{\includegraphics[width=\w\textwidth]{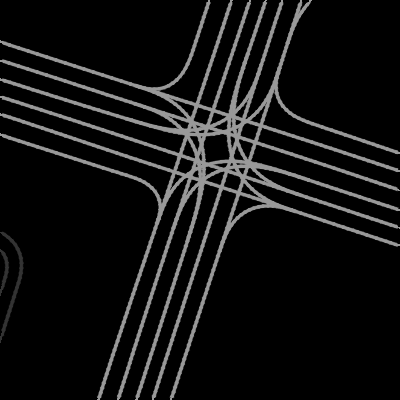}}
\hfill
\subcaptionbox{Route}{\includegraphics[width=\w\textwidth]{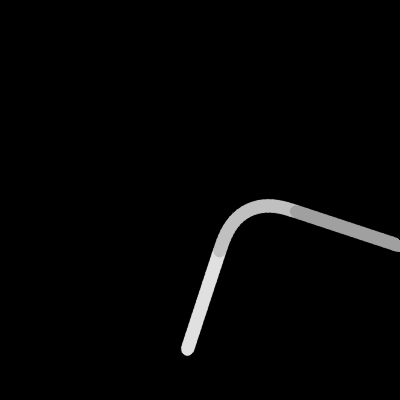}}\\
\subcaptionbox{Current Agent Box}{\includegraphics[width=\w\textwidth]{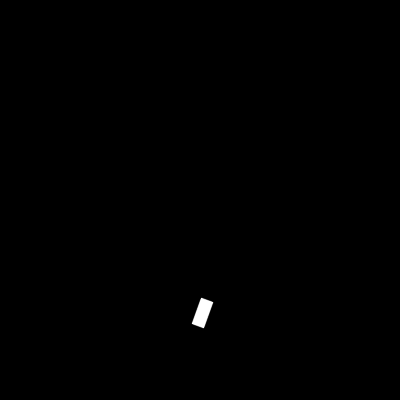}}
\hfill
\subcaptionbox{Dynamic Boxes}{
  \begin{tikzpicture}
        \node[anchor=south west,inner sep=0] (image) at (0,0) {\includegraphics[width=\w\textwidth,cframe=white]{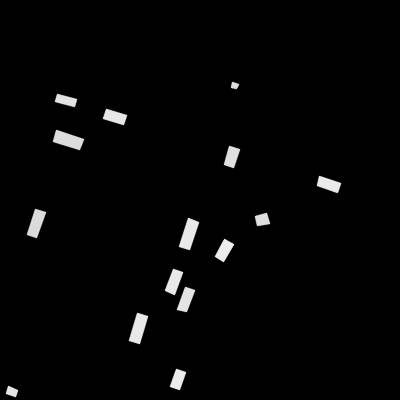}};
        \begin{scope}[x={(image.south east)},y={(image.north west)}]
            \node[anchor=south west,inner sep=0] (image) at (0.025,-0.025) {\includegraphics[width=\w\textwidth,cframe=white]{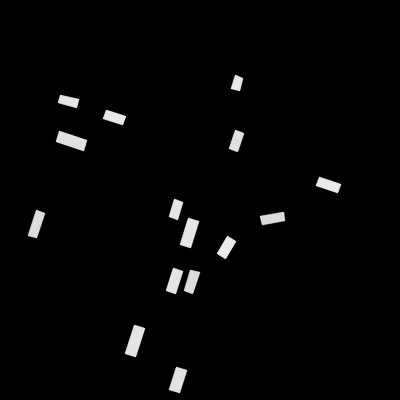}};
            \node[anchor=south west,inner sep=0] (image) at (0.05,-0.05) {\includegraphics[width=\w\textwidth,cframe=white]{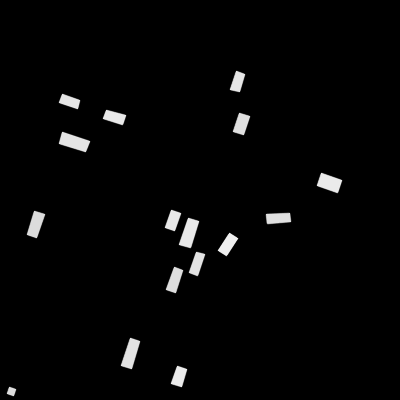}};
            \node[anchor=south west,inner sep=0] (image) at (0.075,-0.075) {\includegraphics[width=\w\textwidth,cframe=white]{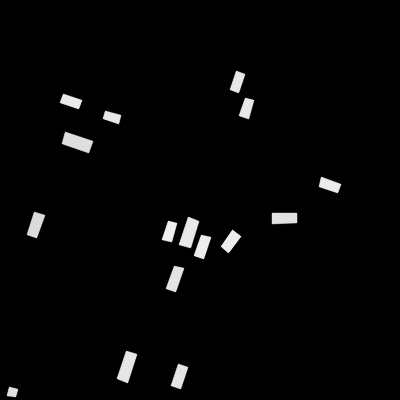}};
            \node[anchor=south west,inner sep=0] (image) at (0.1,-0.1) {\includegraphics[width=\w\textwidth,cframe=white]{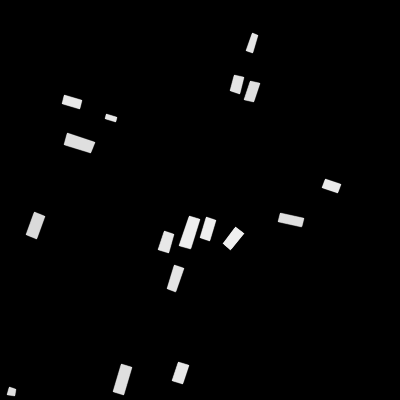}};
        \end{scope}
  \end{tikzpicture}
}
\hfill
\subcaptionbox{Past Agent Poses}{
  \begin{tikzpicture}
      \node[anchor=south west,inner sep=0] (image) at (0,0) {\includegraphics[width=\w\textwidth]{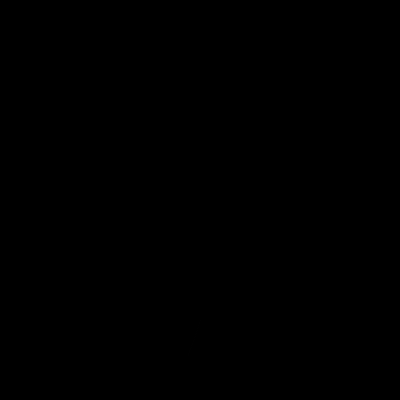}};
      \begin{scope}[x={(image.south east)},y={(image.north west)}]
      \def\x{{0.50001,0.498467,0.497051,0.495664,0.494307,0.492969,0.491641,0.490303,0.488975,0.487647,0.486387,0.485166,0.483936,0.482686,0.481436,0.480195,0.478975,0.477769,0.476592,0.475449,0.474414,0.473521,0.472788,0.472275,0.471982,0.471851,0.471855,0.471851,0.47189,0.47186,0.471738,0.471558,0.471338,0.471069,0.470811,0.470547,0.470264,0.469961,0.469648,0.469307,0.468926}}
      \def\y{{0.79999,0.804443,0.808555,0.812573,0.816504,0.8204,0.824268,0.828125,0.831943,0.835752,0.839375,0.842861,0.846353,0.849873,0.853379,0.856826,0.860195,0.863491,0.866675,0.869756,0.872563,0.874966,0.876943,0.878345,0.879141,0.879365,0.879365,0.879346,0.879238,0.879326,0.879653,0.880137,0.880742,0.881455,0.882153,0.882876,0.883643,0.884453,0.885312,0.886235,0.887271}}
      \foreach \i in {0,...,9}
          \pgfmathparse{\x[\i*2]}%
          \let\p\pgfmathresult
          \pgfmathparse{\y[\i*2]}%
          \let\q\pgfmathresult
          \filldraw[white] (\p, 1.0-\q) circle (0.1pt);
      \end{scope}
  \end{tikzpicture}
}
\hfill
\subcaptionbox{Future Agent Poses}{
  \begin{tikzpicture}
      \node[anchor=south west,inner sep=0] (image) at (0,0) {\includegraphics[width=\w\textwidth]{figures/input_output/rel_pose_image.png}};
      \begin{scope}[x={(image.south east)},y={(image.north west)}]
      \def\x{{0.501704,0.503594,0.505703,0.508047,0.510605,0.513408,0.516406,0.519605,0.522988,0.526543}}
      \def\y{{0.795093,0.789639,0.783623,0.777017,0.769834,0.762095,0.753867,0.745195,0.736162,0.726763}}
      \foreach \i in {0,...,4}
          \pgfmathparse{\x[\i*2]}%
          \let\p\pgfmathresult
          \pgfmathparse{\y[\i*2]}%
          \let\q\pgfmathresult
          \filldraw[green] (\p, 1.0-\q) circle (0.1pt);
      \end{scope}
  \end{tikzpicture}
}
\caption{Driving model inputs (a-g) and output (h).}
\label{fig:io}
\end{figure*}

\subsection{Input Output Representation}
We begin by describing our top-down input representation that the network
will process to output
a drivable trajectory. At any time $t$, our
agent (or vehicle) may be represented in a top-down coordinate system
by $\vec{p}_{t},\theta_{t},s_{t}$, where $\vec{p}_{t}=(x_{t},y_{t})$
denotes the agent's location or pose, $\theta_{t}$ denotes the heading
or orientation, and $s_{t}$ denotes the speed. The top-down coordinate
system is picked such that our agent's pose $\vec{p}_{0}$ at the
current time $t=0$\textsubscript{} is always at a fixed location $(u_0, v_0)$
within the image. For data augmentation purposes during training, the orientation
of the coordinate system is randomly picked for each training example to be within
an angular range of $\theta_{0}\pm\Delta$, where $\theta_{0}$ denotes the heading or
orientation of our agent at time $t=0$. The top-down view is represented by a
set of images of size $W \times H$ pixels, at a ground sampling resolution of
$\phi$ meters/pixel. Note that as the agent moves, this view of the environment moves with it
so the agent always sees a fixed forward range, $R_{forward} = (H - v_0)\phi$ of the world -- similar to having
an agent with sensors that see only up to $R_{forward}$ meters forward.

As shown in \cref{fig:io}, the input to our model consists of several
images of size $W \times H$ pixels rendered into this top-down coordinate system.
(a) Roadmap: a color (3-channel) image with a rendering of various map
features such as lanes, stop signs, cross-walks, curbs, etc.
(b) Traffic lights: a temporal sequence of grayscale images where each frame of
the sequence represents the known state of the traffic lights at each past timestep.
Within each frame, we color each lane center by a gray level with the brightest level for red
lights, intermediate gray level for yellow lights, and a darker level
for green or unknown lights\footnote{We employ an indexed representation for
roadmap and traffic lights channels to reduce the number of input channels, and
to allow extensibility of the input representation to express more roadmap features
or more traffic light states without changing the model architecture.}.
(c) Speed limit: a single channel image with lane centers colored in proportion to their known speed
limit.
(d) Route: the intended route along which we wish to
drive, generated by a router (think of a Google Maps-style route).
(e) Current agent box: this shows our agent's full bounding box at
the current timestep $t=0$.
(f) Dynamic objects in the environment: a temporal sequence of images showing all the potential dynamic
objects (vehicles, cyclists, pedestrians) rendered as oriented boxes.
(g) Past agent poses: the past poses of our agent are rendered into
a single grayscale image as a trail of points.

\def\w{1.0}
\begin{figure}[!t]
\centering \includegraphics[width=\w\textwidth]{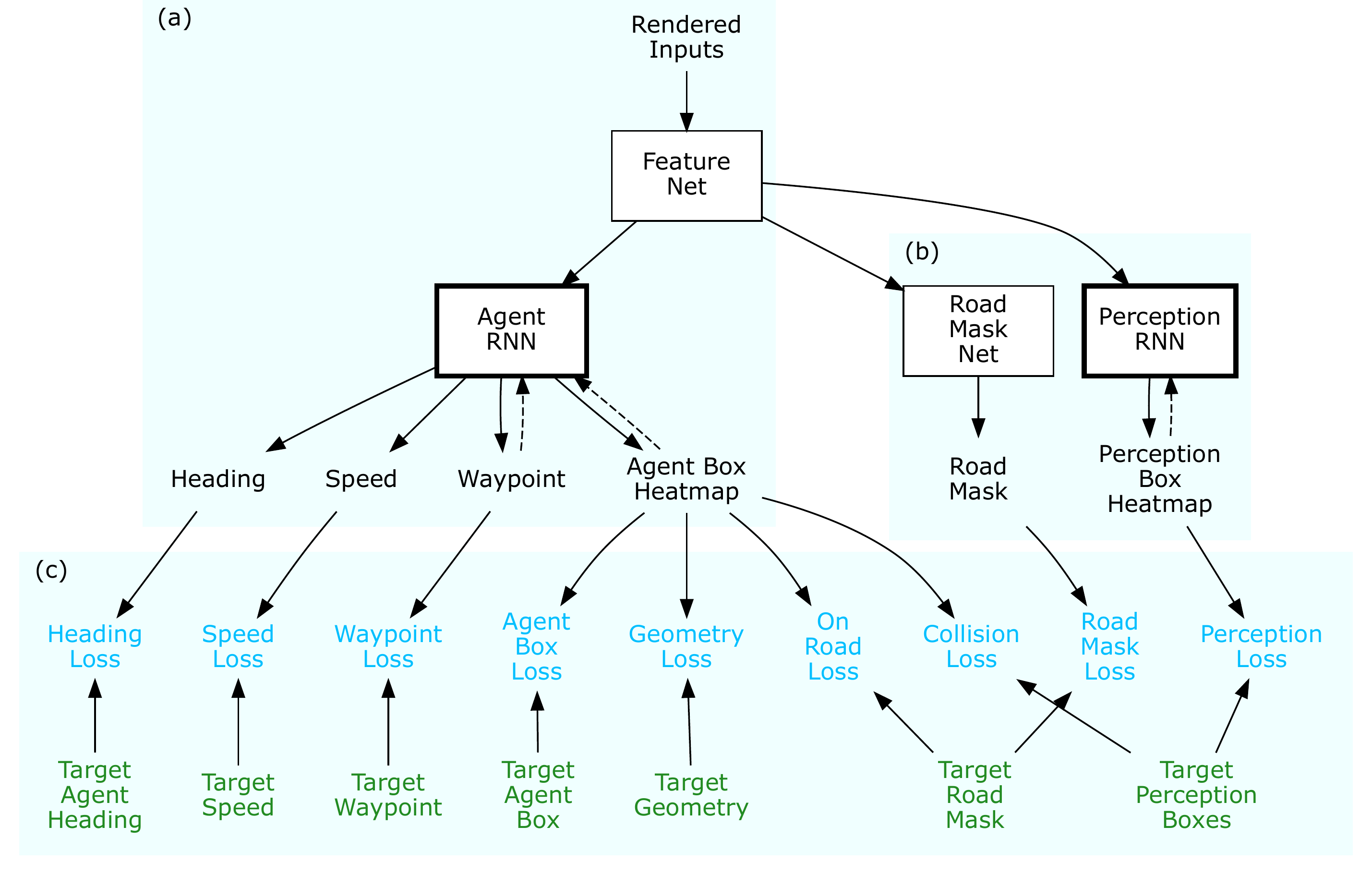}
\caption{Training the driving model. (a) The core {\em ChauffeurNet} model with a FeatureNet and
an AgentRNN, (b) Co-trained road mask prediction net and PerceptionRNN,
and (c) Training losses are shown in blue, and the green labels depict
the ground-truth data. The dashed arrows represent the recurrent feedback of
predictions from one iteration to the next. }
\label{fig:arch}
\end{figure}

We use a fixed-time sampling of $\delta t$ to sample any past or
future temporal information, such as the traffic light state or dynamic
object states in the above inputs. The traffic lights and dynamic
objects are sampled over the past $T_{scene}$ seconds, while the
past agent poses are sampled over a potentially longer interval of
$T_{pose}$ seconds. This simple input representation, particularly
the box representation of other dynamic objects, makes it easy to
generate input data from simulation or create it from real-sensor
logs using a standard perception system that detects and tracks objects.
This enables testing and validation of models in closed-loop simulations
before running them on a real car. This also allows
the same model to be improved using simulated data to adequately explore
rare situations such as collisions for which real-world data might
be difficult to obtain. Using a top-down 2D view also means efficient
convolutional inputs, and allows flexibility to represent metadata and spatial
relationships in a human-readable format. Papers on testing frameworks such as \cite{tian2018deeptest},
\cite{pei2017deepxplore} show the brittleness of using raw sensor
data (such as camera images or lidar point clouds) for learning to
drive, and reinforce the approach of using an intermediate input representation.

If $I$ denotes the set of all the inputs enumerated above, then the
{\em ChauffeurNet} model recurrently predicts future poses of our agent conditioned on
these input images $I$ as shown by the green dots in \cref{fig:io}(h).
\begin{align}
\vec{p}_{t+\delta t}=\textrm{ChauffeurNet}(I,\vec{p}_{t})\label{eq:RNNiter}
\end{align}

In \cref{eq:RNNiter}, current pose $\vec{p}_{0}$ is a known part
of the input, and then the ChauffeurNet performs N iterations and outputs a
future trajectory$\{\vec{p}_{\delta t},\vec{p}_{2\delta t},...,\vec{p}_{N\delta t}\}$
along with other properties such as future speeds. This trajectory
can be fed to a controls optimizer that computes detailed driving
control (such as steering and braking commands) within the specific
constraints imposed by the dynamics of the vehicle to be driven. Different
types of vehicles may possibly utilize different control outputs to
achieve the same driving trajectory, which argues against training
a network to directly output low-level steering and acceleration control.
Note, however, that having intermediate representations like ours does not preclude
end-to-end optimization from sensors to controls.

\subsection{Model Design}

\def\w{0.7}
\def\h{0.6}
\begin{figure}[!t]
\centering \subcaptionbox{\label{fig:iteration:full}}{\includegraphics[width=\w\textwidth]{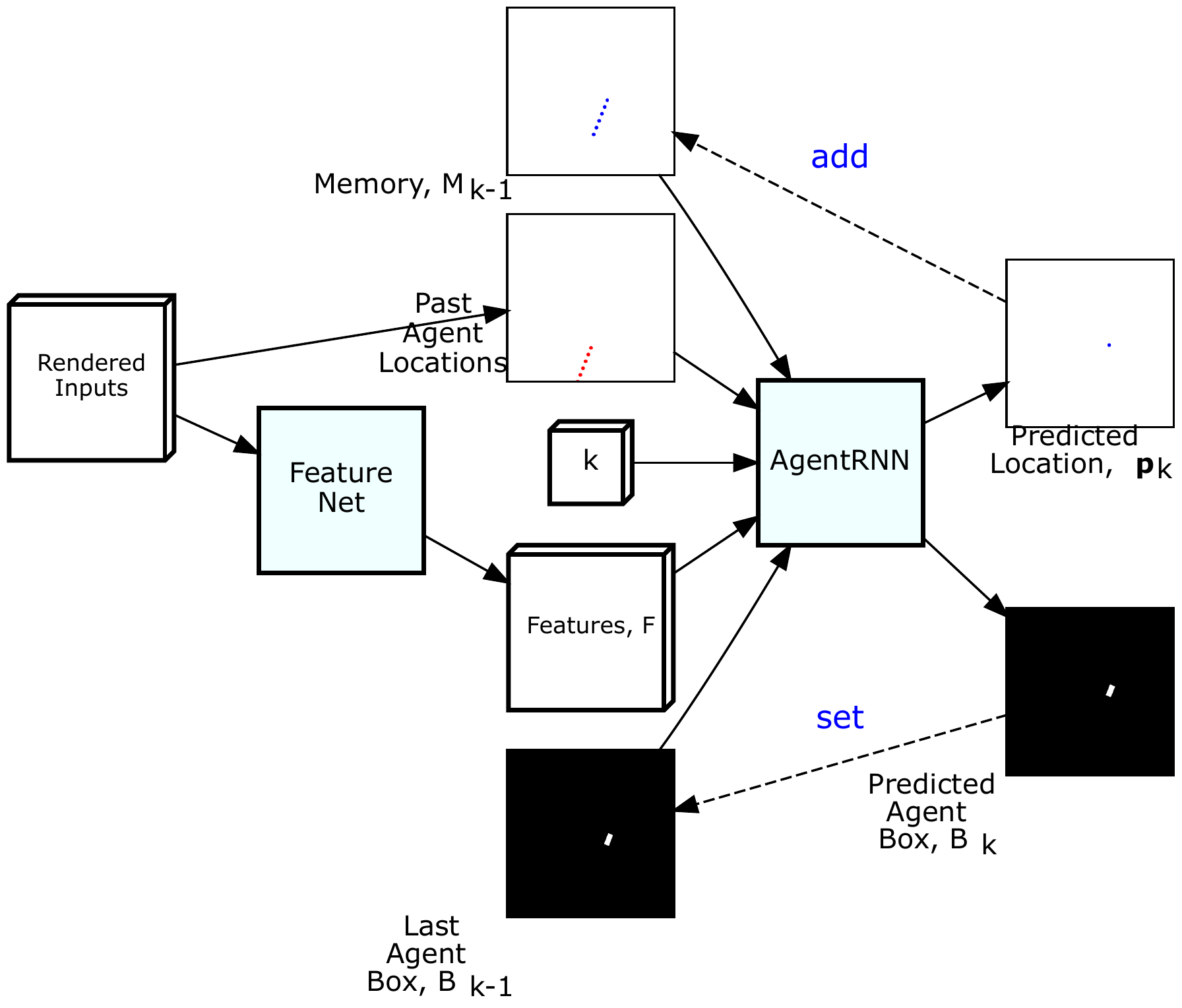}}
\quad\quad\quad\subcaptionbox{\label{fig:iteration:memory}}{\includegraphics[height=\h\textwidth]{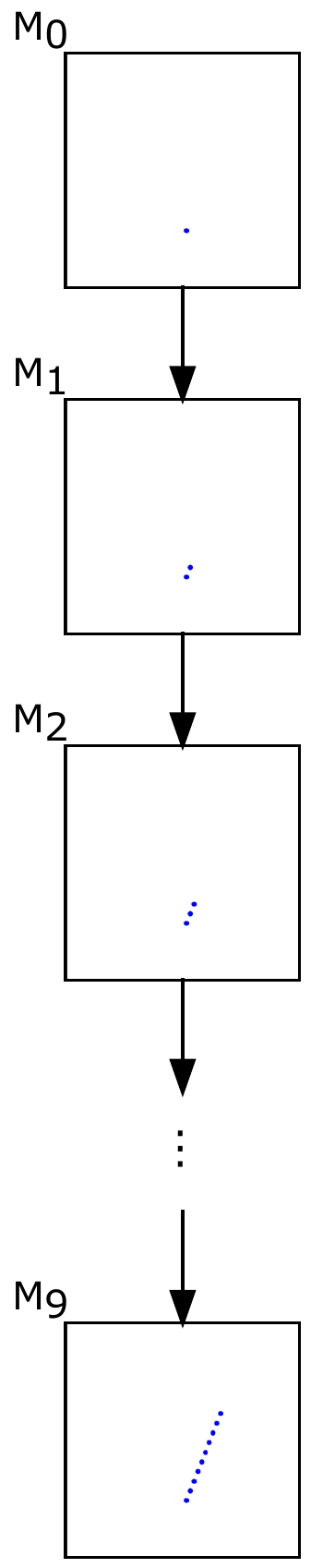}}
\caption{(a) Schematic of ChauffeurNet. (b) Memory updates over multiple iterations.}
\label{fig:iteration}
\end{figure}

Broadly, the driving model is composed of several parts as shown in
\cref{fig:arch}. The main {\em ChauffeurNet} model shown in part (a) of the figure
consists of a convolutional feature network ({\em FeatureNet})
that consumes the input data to create a digested contextual feature
representation that is shared by the other networks. These features
are consumed by a recurrent agent network ({\em AgentRNN}) that
iteratively predicts successive points in the driving trajectory.
Each point at time $t$ in the trajectory is characterized by its location
$\vec{p}_{t}=(x_{t},y_{t})$, heading $\theta_{t}$ and speed $s_{t}$. The {\em AgentRNN} also
predicts the bounding box of the vehicle as a spatial heatmap at each
future timestep. In part (b) of the figure, we see that two other
networks are co-trained using the same feature representation as an
input. The Road Mask Network predicts the drivable areas of the field
of view (on-road vs. off-road), while the recurrent perception network
({\em PerceptionRNN}) iteratively predicts a spatial heatmap for
each timestep showing the future location of every other agent in
the scene. We believe that doing well on these additional tasks
using the same shared features as the main task improves generalization
on the main task. \cref{fig:arch}(c) shows the various losses used in training
the model, which we will discuss in detail below.

\cref{fig:iteration} illustrates the ChauffeurNet model in more detail. The rendered inputs
shown in \cref{fig:io} are fed to a large-receptive field convolutional
{\em FeatureNet} with skip connections, which outputs features
$F$ that capture the environmental context and the intent. These
features are fed to the {\em AgentRNN} which predicts the next
point $\vec{p}_{k}$ on the driving trajectory, and the agent bounding
box heatmap $B_{k}$, conditioned on the features $F$ from the FeatureNet,
the iteration number $k \in \{1,\ldots,N\}$, the memory $M_{k-1}$ of past predictions
from the {\em AgentRNN}, and the agent bounding box heatmap $B_{k-1}$
predicted in the previous iteration.
\begin{align}
\vec{p}_{k},B_{k}=\textrm{AgentRNN}(k,F,M_{k-1},B_{k-1})\label{eq:AgentRNNiter}
\end{align}

The memory $M_{k}$ is an additive memory consisting of a single channel
image. At iteration $k$ of the {\em AgentRNN}, the memory is incremented
by 1 at the location $\vec{p}_{k}$ predicted by the {\em AgentRNN},
and this memory is then fed to the next iteration. The {\em AgentRNN}
outputs a heatmap image over the next pose of the agent,
and we use the arg-max operation to obtain the coarse pose prediction
$\vec{p}_{k}$ from this heatmap. The {\em AgentRNN} then employs
a shallow convolutional meta-prediction network with a fully-connected layer that predicts a sub-pixel refinement of the
pose $\delta\vec{p}_{k}$ and also estimates the heading $\theta_{k}$ and the speed $s_{k}$. Note
that the {\em AgentRNN} is unrolled at training time for a
fixed number of iterations, and the losses described below are summed
together over the unrolled iterations. This is possible because of the
non-traditional RNN design where we employ an explicitly crafted memory model
instead of a learned memory.

\subsection{System Architecture}
\begin{figure*}[!t]
\centering \includegraphics[width=1\textwidth]{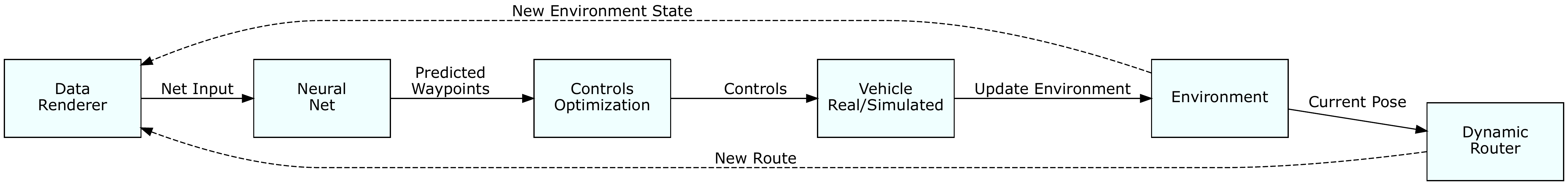} \caption{Software architecture for the end-to-end driving pipeline.}
\label{fig:sw}
\end{figure*}

\cref{fig:sw} shows a system level overview of how the neural net
is used within the self-driving system. At each time, the updated
state of our agent and the environment is obtained via a perception
system that processes sensory output from the real-world or from a
simulation environment as the case may be. The intended route is obtained
from the router, and is updated dynamically conditioned on whether
our agent was able to execute past intents or not. The environment
information is rendered into the input images described in \cref{fig:io}
and given to the RNN which then outputs a future trajectory. This
is fed to a controls optimizer that outputs the low-level control
signals that drive the vehicle (in the real world or in simulation).

\section{Imitating the Expert}
In this section, we first show how to train the model above to imitate the expert.
\subsection{Imitation Losses\label{sec:imitation_loss}}
\subsubsection{Agent Position, Heading and Box Prediction}

The {\em AgentRNN} produces three outputs at each iteration $k$:
a probability distribution $P_{k}(x,y)$ over the spatial coordinates
of the predicted waypoint obtained after a spatial {\em softmax},
a heatmap of the predicted agent box at that timestep $B_{k}(x,y)$
obtained after a per-pixel {\em sigmoid} activation that represents
the probability that the agent occupies a particular pixel, and a
regressed box heading output $\theta_{k}$. Given ground-truth data
for the above predicted quantities, we can define the corresponding
losses for each iteration as:
\begin{align}
\mathcal{L}_{p} & =\mathcal{H}(P_{k},P_{k}^{gt})\\
\mathcal{L}_{B} & =\frac{1}{WH}\sum_{x}{\sum_{y}{\mathcal{H}(B_{k}(x,y),B_{k}^{gt}(x,y))}}\\
\mathcal{L}_{\theta} & =\norm{\theta_{k}-\theta_{k}^{gt}}_{1}
\end{align}
where the superscript {\em gt} denotes the corresponding ground-truth
values, and $\mathcal{H}(a,b)$ is the cross-entropy function. Note that
$P_{k}^{gt}$ is a binary image with only the pixel at the ground-truth target
coordinate $\floor{\vec{p}_{k}^{gt}}$ set to one.

\subsubsection{Agent Meta Prediction}

The meta prediction network performs regression on the features to generate
a sub-pixel refinement $\delta\vec{p}_{k}$ of the coarse waypoint
prediction as well as a speed estimate $s_k$ at each iteration. We employ
$L_{1}$ loss for both of these outputs:
\begin{align}
\mathcal{L}_{p-subpixel} & =\norm{\delta\vec{p}_{k}-\delta\vec{p}_{k}^{gt}}_{1}\\
\mathcal{L}_{speed} & =\norm{s_{k}-s_{k}^{gt}}_{1}
\end{align}
where $\delta\vec{p}_{k}^{gt} = \vec{p}_{k}^{gt} - \floor{\vec{p}_{k}^{gt}}$ is
the fractional part of the ground-truth pose coordinates.
\subsection{Past Motion Dropout \label{sec:dropout}}

During training, the model is provided the past motion history as
one of the inputs (\cref{fig:io}(g)). Since the past motion history
during training is from an expert demonstration, the net can learn
to ``cheat'' by just extrapolating from the past rather than finding
the underlying causes of the behavior. During closed-loop inference,
this breaks down because the past history is from the net's own past
predictions. For example, such a trained net may learn to only stop
for a stop sign if it sees a deceleration in the past history, and
will therefore never stop for a stop sign during closed-loop inference.
To address this, we introduce a dropout on the past pose history, where for $50\%$ of
the examples, we keep only the current position $(u_0, v_0)$ of the agent in the
past agent poses channel of the input data. This forces the net to look at other cues in the environment
to explain the future motion profile in the training example.

\section{Beyond Pure Imitation}

In this section, we go beyond vanilla cloning of the expert's demonstrations in
order to teach the model to arrest drift and avoid bad behavior such as
collisions and off-road driving by synthesizing variations of the expert's behavior.

\subsection{Synthesizing Perturbations \label{sec:perturb}}

\def\w{0.45}
\begin{figure*}[t]
\centering \subcaptionbox{Original}{\includegraphics[width=\w\textwidth]{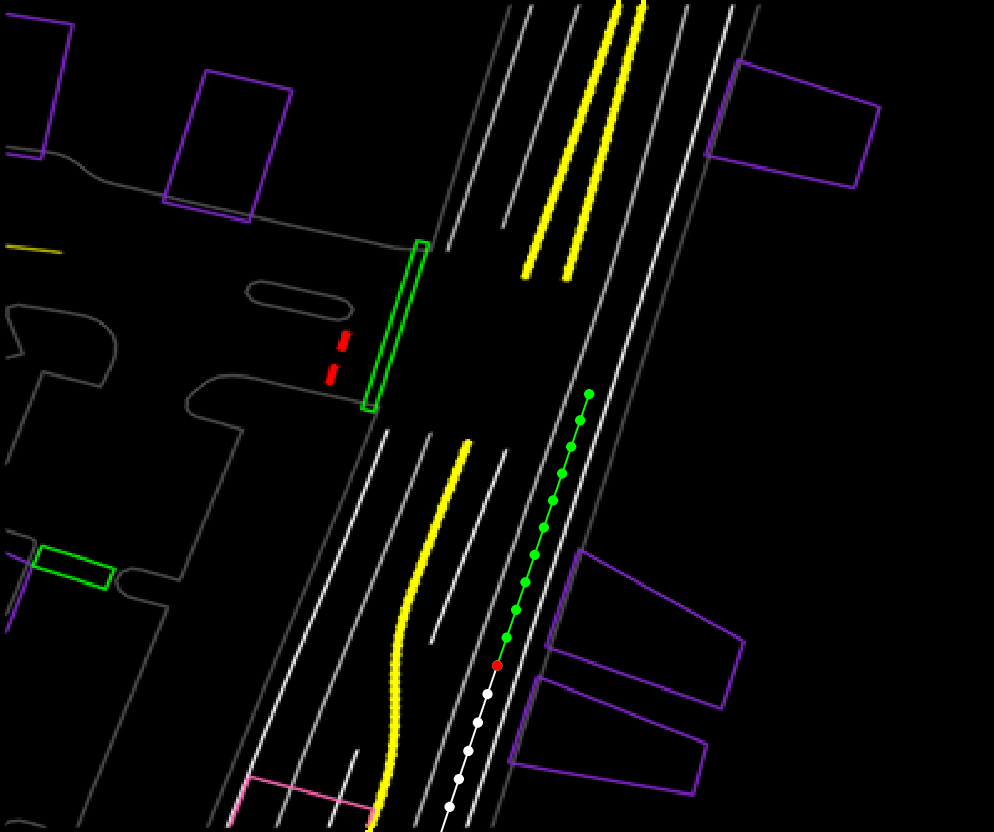}}\quad{}\subcaptionbox{Perturbed}{\includegraphics[width=\w\textwidth]{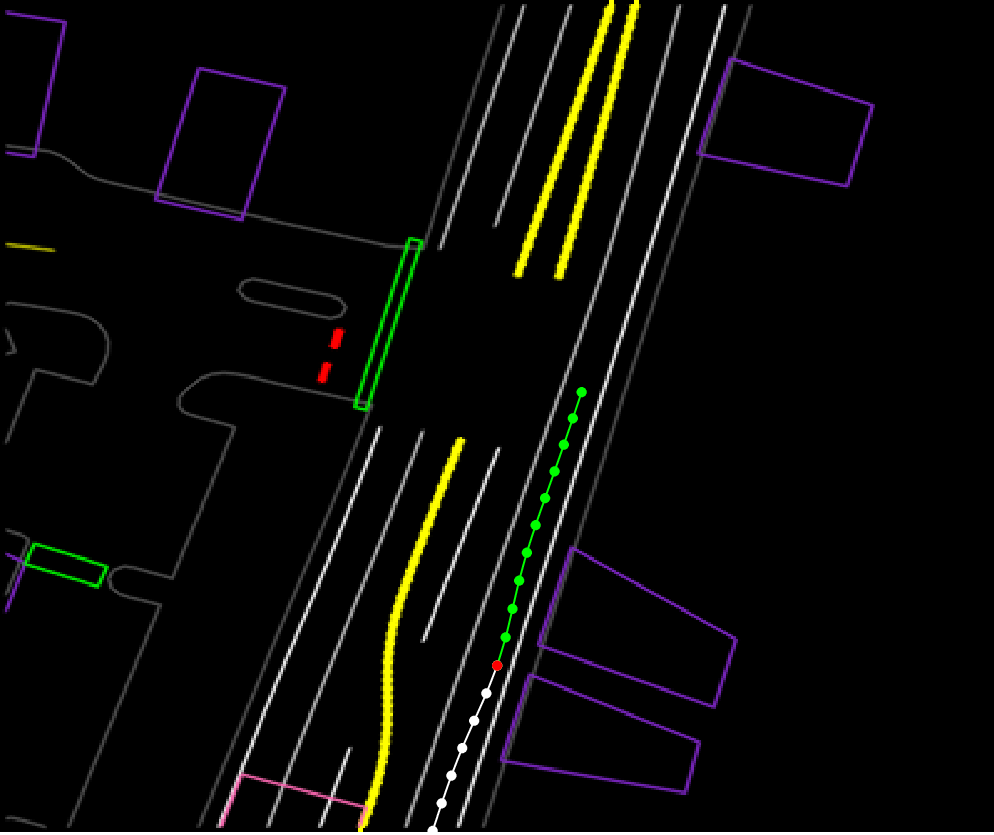}}
\caption{Trajectory Perturbation. (a) An original logged training example where the agent is
driving along the center of the lane. (b) The perturbed example created by perturbing the current agent location (red point) in the original
example away from the lane center and then fitting a new smooth trajectory that  brings the
agent back to the original target location along the lane center.\label{fig:perturb_example}}
\end{figure*}

Running the model as a part of a closed-loop system over time can
cause the input data to deviate from the training distribution. To
prevent this, we train the model by adding some examples with realistic
perturbations to the agent trajectories. The start and end of a trajectory
are kept constant, while a perturbation is applied around the midpoint
and smoothed across the other points. Quantitatively, we jitter the
midpoint pose of the agent uniformly at random in the range $[-0.5,0.5]$
meters in both axes, and perturb the heading by $[-\pi/3,\pi/3]$
radians. We then fit a smooth trajectory to the perturbed point and
the original start and end points. Such training examples bring the
car back to its original trajectory after a perturbation. \cref{fig:perturb_example}
shows an example of perturbing the current agent location (red point) away
from the lane center and the fitted trajectory correctly
bringing it back to the original target location along the lane center. We filter
out some perturbed trajectories that are impractical by thresholding
on maximum curvature. But we do allow the perturbed trajectories to
collide with other agents or drive off-road, because the network can
then experience and avoid such behaviors even though real examples
of these cases are not present in the training data. In training,
we give perturbed examples a weight of $1/10$ relative to the real
examples, to avoid learning a propensity for perturbed driving.

\subsection{Beyond the Imitation Loss\label{sec:beyond_imitation_loss}}
\subsubsection{Collision Loss}

Since our training data does not have any real collisions, the idea of avoiding
collisions is implicit and will not generalize well. To alleviate
this issue, we add a specialized loss that directly measures the overlap
of the predicted agent box $B_{k}$ with the ground-truth boxes of
all the scene objects at each timestep.
\begin{align}
\mathcal{L}_{collision}=\frac{1}{WH}\sum_{x}{\sum_{y}{B_{k}(x,y)\mult Obj_{k}^{gt}(x,y)}}
\end{align}
where $B_{k}$ is the likelihood map for the output agent box prediction, and $Obj_{k}^{gt}$ is a binary mask with
ones at all pixels occupied by other dynamic objects (other vehicles,
pedestrians, etc.) in the scene at timestep $k$. At any time during
training, if the model makes a poor prediction that leads to a collision,
the overlap loss would influence the gradients to correct the mistake.
However, this loss would be effective only during
the initial training rounds when the model hasn't learned to predict close to the
ground-truth locations due to the absence of real collisions in the
ground truth data. This issue is alleviated by the addition of trajectory perturbation
data, where artificial collisions within those examples allow this loss to be effective
throughout training without the need for online exploration like in reinforcement learning settings.

\subsubsection{On Road Loss}

Trajectory perturbations also create synthetic cases where the car veers off the
road or climbs a curb or median because of the perturbation. To train the network to
avoid hitting such hard road edges, we add a specialized
loss that measures overlap of the predicted agent box $B_{k}$ in
each timestep with a binary mask $Road^{gt}$ denoting the road and
non-road regions within the field-of-view.
\begin{align}
\mathcal{L}_{onroad}=\frac{1}{WH}\sum_{x}{\sum_{y}{B_{k}(x,y)\mult(1-Road^{gt}(x,y))}}
\end{align}

\subsubsection{Geometry Loss}

We would like to explicitly constrain the agent to follow the target
geometry independent of the speed profile. We model this target geometry
by fitting a smooth curve to the target waypoints and rendering this
curve as a binary image in the top-down coordinate system. The thickness
of this curve is set to be equal to the width of the agent. We express
this loss similar to the collision loss by measuring the overlap of
the predicted agent box with the binary target geometry image $Geom^{gt}$.
Any portion of the box that does not overlap with the target geometry
curve is added as a penalty to the loss function.
\begin{align}
\mathcal{L}_{geom}=\frac{1}{WH}\sum_{x}{\sum_{y}{B_{k}(x,y)\mult(1-Geom{}^{gt}(x,y))}}
\end{align}

\begin{figure*}[!t]
\centering
\subcaptionbox{Flattened Inputs}{\includegraphics[width=0.24\textwidth]{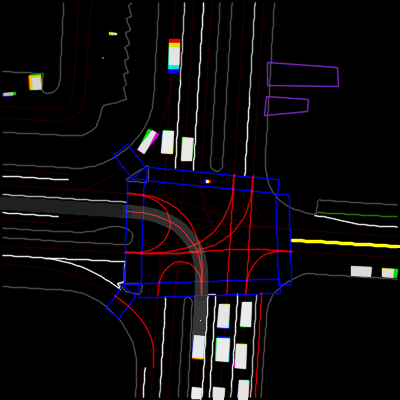}}
\subcaptionbox{Target Road Mask}{\includegraphics[width=0.24\textwidth]{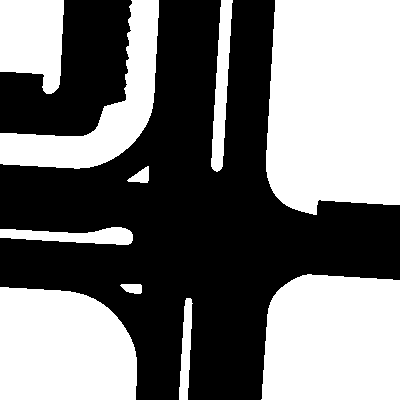}}
\subcaptionbox{Pred Road Mask Logits}{\includegraphics[width=0.24\textwidth]{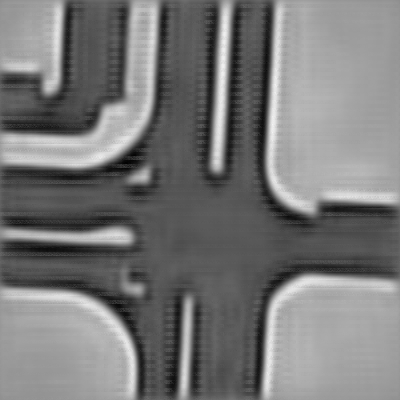}}
\subcaptionbox{Pred Vehicles Logits}{\includegraphics[width=0.24\textwidth]{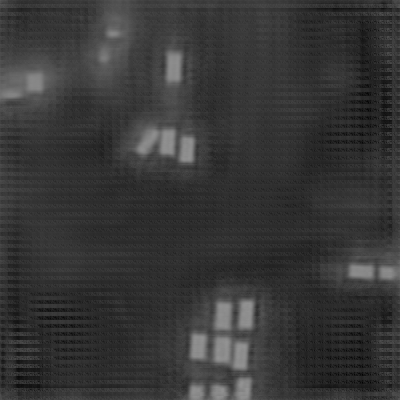}}
\subcaptionbox{Agent Pose Logits}{\includegraphics[width=0.24\textwidth]{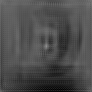}}
\subcaptionbox{Collision Loss}{\includegraphics[width=0.24\textwidth]{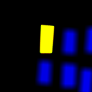}}
\subcaptionbox{On Road Loss}{\includegraphics[width=0.24\textwidth]{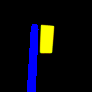}}
\subcaptionbox{Geometry Loss}{\includegraphics[width=0.24\textwidth]{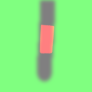}}
\caption{Visualization of predictions and loss functions on an example input. The top row is at the
input resolution, while the bottom row shows a zoomed-in view around the current agent location.}
\label{fig:example}
\end{figure*}

\subsubsection{Auxiliary Losses}
Similar to our own agent's trajectory, the motion of other agents
may also be predicted by a recurrent network. Correspondingly, we
add a recurrent perception network {\em PerceptionRNN} that uses
as input the shared features $F$ created by the {\em FeatureNet}
and its own predictions $Obj_{k-1}$ from the previous iteration,
and predicts a heatmap $Obj_{k}$ at each iteration. $Obj_{k}(x,y)$
denotes the probability that location $(x,y)$ is occupied by a dynamic
object at time $k$. For iteration $k=0$, the PerceptionRNN is
fed the ground truth objects at the current time.
\begin{align}
\mathcal{L}_{objects} & =\frac{1}{WH}\sum_{x}{\sum_{y}{\mathcal{H}(Obj_{k}(x,y),Obj_{k}^{gt}(x,y))}}
\end{align}
Co-training a {\em PerceptionRNN} to predict the future of other agents by sharing the same
feature representation $F$ used by the {\em PerceptionRNN} is likely to induce the feature
network to learn better features that are suited to both tasks. Several examples of predicted
trajectories from {\em PerceptionRNN} on logged data are shown on our website \href{\site}{here}.

We also co-train to predict a binary road/non-road mask by adding a small
network of convolutional layers to the output of the feature net $F$. We add a cross-entropy loss to the
predicted road mask output $Road(x,y)$ which compares it to the ground-truth road mask $Road^{gt}$.
\begin{align}
\mathcal{L}_{road} & =\frac{1}{WH}\sum_{x}{\sum_{y}{\mathcal{H}(Road(x,y),Road{}^{gt}(x,y))}}
\end{align}

\cref{fig:example} shows some of the predictions and losses for a single example
processed through the model.

\subsection{Imitation Dropout \label{sec:sample_loss}}

Overall, our losses may be grouped into two sub-groups, the imitation losses:
\begin{align}
\Lagent=\{\Loss_{p},\Loss_{B},\Loss_{\theta},\Loss_{p-subpixel},\Loss_{speed}\}
\end{align}

and the environment losses:
\begin{align}
\Lenv=\{\Loss_{collision},\Loss_{onroad},\Loss_{geom},\Loss_{objects},\Loss_{road}\}
\end{align}

The imitation losses cause the model to imitate the expert's demonstrations,
while the environment losses discourage undesirable behavior such
as collisions. To further increase the effectiveness of the environment
losses, we experimented with randomly dropping out the imitation losses
for a random subset of training examples. We refer to this as ``imitation
dropout''. In the experiments, we show that imitation dropout yields
a better driving model than simply under-weighting the imitation losses.
During imitation dropout, the weight on the imitation losses $\wagent$
is randomly chosen to be either 0 or 1 with a certain probability
for each training example. The overall loss is given by:
\begin{align}
\Loss=\wagent\sum_{\ell\in\Lagent}{\ell}+\wenv\sum_{\ell\in\Lenv}{\ell}
\end{align}

\section{Experiments}
\subsection{Data}

\begin{table}[t]
\centering
\begin{tabular}{cccccccccc}
\hline
$T_{scene}$ & $T_{pose}$ & $\delta t$ & $N$ & $\Delta$ & $W$ & $H$ & $u_0$ & $v_0$ & $\phi$\\
1.0 s & 8.0 s & 0.2s & 10 & $25^\circ$ & 400 px & 400 px & 200 px & 320 px & 0.2 m/px \\
\hline
\end{tabular}
\caption{Parameter values for the experiments in this paper.}
\label{table:params}
\end{table}

\begin{table}[t]
\centering
\begin{tabular}{p{0.13\textwidth}p{0.13\textwidth}p{0.23\textwidth}p{0.27\textwidth}p{0.1\textwidth}}
\hline
Rendering & FeatureNet & AgentRNN (N=10) & PerceptionRNN (N=10) & \textbf{Overall} \\
8 ms & 6.5 ms & 145 ms & 35 ms & \textbf{160 ms}\\
\hline
\end{tabular}
\caption{Run-time performance on NVIDIA Tesla P100 GPU.}
\label{table:runtime}
\end{table}

The training data to train our model was obtained by randomly sampling
segments of real-world expert driving and removing segments where the car was
stationary for long periods of time. Our input field of view is $80m\times80m$ ($W\phi = 80$) and with the
agent positioned at $(u_0, v_0)$, we get an effective forward sensing range of $R_{forward} = 64m$.
Therefore, for the experiments in this work we also removed any segments of highway driving given
the longer sensing range requirement that entails. Our dataset contains approximately
26 million examples which amount to about 60 days of continuous driving.
As discussed in \cref{sec:arch}, the vertical-axis of the
top-down coordinate system for each training example is randomly oriented
within a range of $\Delta=\pm25^{\circ}$ of our agent's current heading,
in order to avoid a bias for driving along the vertical axis. The rendering
orientation is set to the agent heading ($\Delta=0$) during inference. Data
about the prior map of the environment (roadmap) and the speed-limits along the
lanes is collected apriori. For the dynamic scene entities like objects and
traffic-lights, we employ a separate perception system based on laser and camera
data similar to existing works in the literature (\cite{yang2018hdnet, fairfield2011traffic}).
\cref{table:params} lists the parameter values used for all the experiments in this
paper. The model runs on a NVidia Tesla P100 GPU in 160ms with the detailed breakdown in \cref{table:runtime}.

\subsection{Models}
We train and test not only our final model, but a sequence of models that introduce the
ingredients we describe one by one on top of behavior cloning. We start with $\M{0}$,
which does behavior cloning with past motion dropout to prevent using the history to cheat.
$\M{1}$ adds perturbations without modifying the losses. $\M{2}$ further adds our environment losses
$\Lenv$ in \cref{sec:beyond_imitation_loss}. $\M{3}$ and $\M{4}$ address the fact that we do not want to imitate
bad behavior -- $\M{3}$ is a baseline approach, where we simply decrease the weight on the imitation loss,
while $\M{4}$ uses our imitation dropout approach with a dropout probability of $0.5$.
\cref{table:model_config} lists the configuration for each of these models.

\subsection{Closed Loop Evaluation}
To evaluate our learned model on a specific scenario, we replay the segment through the simulation until a buffer period of
$\textrm{max}(T_{pose}, T_{scene})$ has passed. This allows us to generate the first rendered
snapshot of the model input using all the replayed messages until now. The model is evaluated
on this input, and the fitted controls are passed to the vehicle simulator
that emulates the dynamics of the vehicle thus moving the simulated agent to its next pose. At this point,
the simulated pose might be different from the logged pose, but our input representation allows us
to correctly render the new input for the model relative to the new pose. This
process is repeated until the end of the segment, and we evaluate scenario specific metrics like
stopping for a stop-sign, collision with another vehicle etc. during the simulation. Since the model is being
used to drive the agent forward, this is a \emph{closed-loop} evaluation setup.

\subsubsection{Model Ablation Tests}
Here, we present results from experiments using the various models in the closed-loop simulation setup.
We first evaluated all the models on simple situations such as stopping for stop-signs and red traffic lights,
and lane following along straight and curved roads by creating 20 scenarios for each situation, and found that
\emph{all the models worked well in these simple cases}. Therefore, we will focus below on specific complex
situations that highlight the differences between these models.

\begin{table}[t]
\centering
\begin{tabular}{p{0.1\textwidth}p{0.375\textwidth}p{0.2\textwidth}p{0.2\textwidth}}
\hline
\textbf{Model} & \textbf{Description} & $\wagent$ & $\wenv$ \\
$\M{0}$ &  Imitation with Past Dropout & 1.0 &  0.0  \\
$\M{1}$ &  $\M{0}$ + Traj Perturbation & 1.0 & 0.0 \\
$\M{2}$ &  $\M{1}$ + Environment Losses & 1.0 & 1.0 \\
$\M{3}$ &  $\M{2}$ with less imitation  & 0.5 & 1.0 \\
$\M{4}$ &  $\M{2}$ with Imitation Dropout & \multicolumn{2}{c}{\em{Dropout probability = 0.5 (see \cref{sec:sample_loss}).}} \\
\hline
\end{tabular}
\caption{Model configuration for the model ablation tests.}
\label{table:model_config}
\end{table}

\definecolor{wblue}{RGB}{0,120,255}
\definecolor{wgreen}{RGB}{0, 232, 157}
\definecolor{wgray}{RGB}{148, 154, 159}
\def\r{0.8}
\def\w{0.16}

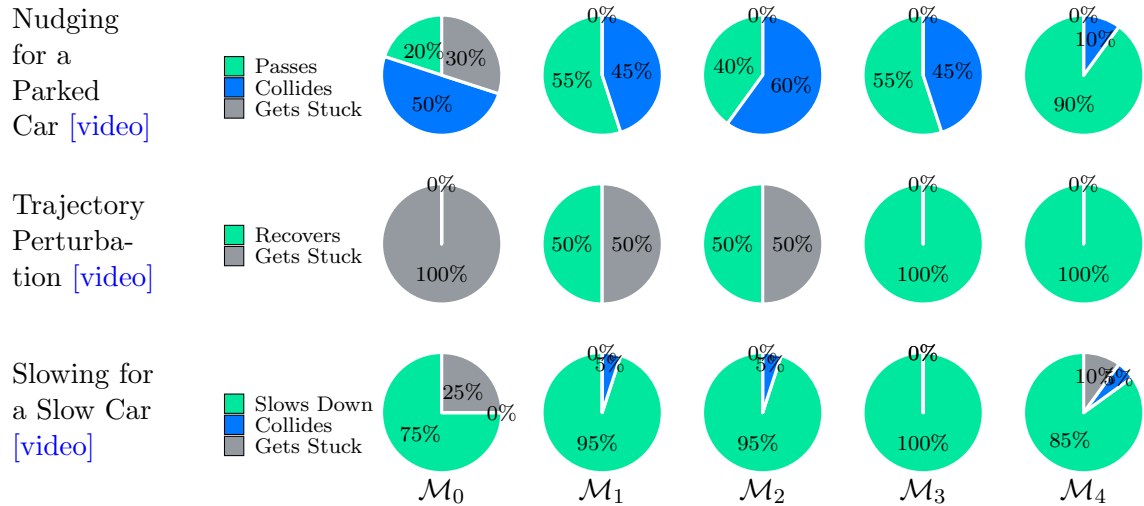
\begin{figure*}[!tbp]
\centering
\begin{tikzpicture}
[
pie chart,
slice type={Passes}{wgreen},
slice type={Collides}{wblue},
slice type={Stuck}{wgray},
pie values/.style={font={\scriptsize}},
scale=\r
]
\node[text width=1.9cm,xshift=-1cm]{Nudging for a Parked Car \href{\site}{[video]}};
\legend[shift={(1.25cm,.5cm)}]{{Passes}/Passes, {Collides}/Collides, {Gets Stuck}/Stuck}
\pie[xshift=1.85in]{}{20/Passes, 50/Collides, 30/Stuck}
\pie[xshift=2.9in]{}{55/Passes, 45/Collides, 0/Stuck}
\pie[xshift=3.95in]{}{40/Passes, 60/Collides, 0/Stuck}
\pie[xshift=5.in]{}{55/Passes, 45/Collides, 0/Stuck}
\pie[xshift=6.05in]{}{90/Passes, 10/Collides, 0/Stuck}
\end{tikzpicture} \\
\begin{tikzpicture}
[
pie chart,
slice type={Recovers}{wgreen},
slice type={Stuck}{wgray},
pie values/.style={font={\scriptsize}},
scale=\r
]
\node[text width=1.9cm,xshift=-1cm]{Trajectory Perturbation \href{\site}{[video]}};
\legend[shift={(1.25cm,0.5cm)}]{{Recovers}/Recovers, {Gets Stuck}/Stuck}
\pie[xshift=1.85in]{}{0/Recovers, 100/Stuck}
\pie[xshift=2.9in]{}{50/Recovers, 50/Stuck}
\pie[xshift=3.95in]{}{50/Recovers, 50/Stuck}
\pie[xshift=5.in]{}{100/Recovers, 0/Stuck}
\pie[xshift=6.05in]{}{100/Recovers, 0/Stuck}
\end{tikzpicture} \\
\begin{tikzpicture}
[
pie chart,
slice type={Slows Down}{wgreen},
slice type={Collides}{wblue},
slice type={Stuck}{wgray},
pie values/.style={font={\scriptsize}},
scale=\r
]
\node[text width=1.9cm,xshift=-1cm]{Slowing for a Slow Car \href{\site}{[video]}};
\legend[shift={(1.25cm,0.5cm)}]{{Slows Down}/Slows Down, {Collides}/Collides, {Gets Stuck}/Stuck}
\pie[xshift=1.85in]{$\M{0}$}{75/Slows Down, 0/Collides, 25/Stuck}
\pie[xshift=2.9in]{$\M{1}$}{95/Slows Down, 5/Collides, 0/Stuck}
\pie[xshift=3.95in]{$\M{2}$}{95/Slows Down, 5/Collides, 0/Stuck}
\pie[xshift=5.0in]{$\M{3}$}{100/Slows Down, 0/Collides, 0/Stuck}
\pie[xshift=6.05in]{$\M{4}$}{85/Slows Down, 5/Collides, 10/Stuck}
\end{tikzpicture} \\
\caption{Model ablation test results on three scenario types.}
\label{fig:pies}
\end{figure*}

\paragraph{Nudging around a parked car.} To set up this scenario, we place the agent at an arbitrary distance from a stop-sign
on an undivided two-way street and then place a parked vehicle on the right shoulder
between the the agent and the stop-sign. We pick 4 separate locations with
both straight and curved roads then vary the starting
speed of the agent between 5 different values to create a total of 20 scenarios.
We then observe if the agent would stop and get stuck behind, collide with
the parked car, or correctly pass around the parked car, and report the aggregate performance
in \cref{fig:pies}(row 1). We find that other than $\M{4}$, all other models cause the agent
to collide with the parked vehicle about half the time. The baseline $\M{0}$ model
can also get stuck behind the parked vehicle in some of the scenarios. The model $\M{4}$ nudges
around the parked vehicle and then brings the agent back to the lane center.
This can be attributed to the model's ability to learn to avoid collisions
and nudge around objects because of training with the collision loss the trajectory perturbation.
Comparing model $\M{3}$ and $\M{4}$, it is apparent that ``imitation dropout'' was more effective
at learning the right behavior than only re-weighting the imitation losses. Note that in this
scenario, we generate several variations by changing the starting speed of the agent relative to the parked car.
This creates situations of increasing difficulty, where the agent approaches the parked car at very high
relative speed and thus does not have enough time to nudge around the car given the dynamic constraints.
A 10\% collision rate for $\M{4}$ is thus not a measure of the absolute performance of the model since
we do not have a perfect driver which could have performed well at all the scenarios here. But in
relative terms, this model performs the best.

\paragraph{Recovering from a trajectory perturbation.} To set up this scenario, we place the agent approaching a curved road and vary the
starting position and the starting speed of the agent to generate a total of 20
scenario variations. Each variation puts the agent at a different amount of offset from the lane center
with a different heading error relative to the lane. We then measure how well the various models are at
recovering from the lane departure.
\cref{fig:pies}(row 2) presents the results aggregated across these scenarios and shows
the contrast between the baseline model $\M{0}$ which is not able to recover in
any of the situations and the models $\M{3}$ and $\M{4}$ which handle all deviations well.
All models trained with the perturbation data are able to handle 50\% of the
scenarios which have a lower starting speed. At a higher starting speed, we believe that
$\M{3}$ and $\M{4}$ do better than $\M{1}$ and $\M{2}$ because they place a higher
emphasis on the imagination losses.

\paragraph{Slowing down for a slow car.} To set up this scenario, we place the agent on a straight road at varying initial speeds
and place another car ahead with a varying but slower constant speed, generating
a total of 20 scenario variations, to evaluate the ability to slow for and then follow the
car ahead. From \cref{fig:pies}(row 3), we see that some models slow down to zero speed and get stuck.
For the variation with the largest relative speed, there isn't enough time for most models
to stop the agent in time, thus leading to a collision. For these cases, model $\M{3}$
which uses imitation loss re-weighting works better than the model $\M{4}$ which uses
imitation dropout. $\M{4}$ has trouble in two situations due to being over aggressive in
trying to maneuver around the slow car and then grazes the left edge of the road. This happens
in the two extreme variations where the relative speed between the two cars is the highest.

\subsubsection{Input Ablation Tests}
With input ablation tests, we want to test the final $\M{4}$ model's ability to identify the correct
causal factors behind specific behaviors, by testing the model's behavior in the
presence or absence of the correct causal factor while holding other conditions
constant. In simulation, we have evaluated our model on 20 scenarios with and
without stop-signs rendered, and 20 scenarios with and without other vehicles in the scene
rendered. The model \href{\site}{exhibits} the correct behavior in all scenarios, thus confirming
that it has learned to respond to the correct features for a stop-sign and a stopped vehicle.

\subsubsection{Logged Data Simulated Driving}
For this evaluation, we take logs from our real-driving test data (separate from our training data),
and use our trained network to drive the car using the vehicle simulator keeping everything else
the same i.e. the dynamic objects, traffic-light states etc. are all kept the same as in the logs.
Some example videos are shown \href{\site}{here} and they illustrate the ability of the
model in dealing with multiple dynamic objects and road controls.

\subsubsection{Real World Driving}
We have also evaluated this model on our self-driving car by replacing the
existing planner module with the learned model $\M{4}$ and have replicated the driving behaviors observed in
simulation. The videos of several of these runs are available \href{\site}{here} and they illustrate not only
the smoothness of the network's driving ability, but also its ability to deal with stop-signs and turns and to
drive for long durations in full closed-loop control without deviating from the trajectory.

\pgfplotstableread[col sep=colon]{
wp:err
waypoint0:0.164275
waypoint1:0.3969
waypoint2:0.661117
waypoint3:0.94868
waypoint4:1.27148
waypoint5:1.64594
waypoint6:2.04987
waypoint7:2.50679
waypoint8:3.01111
waypoint9:3.55167
}\baselineu

\pgfplotstableread[col sep=colon]{
wp:err
waypoint0:0.244298
waypoint1:0.522665
waypoint2:0.800158
waypoint3:1.11986
waypoint4:1.45648
waypoint5:1.82732
waypoint6:2.22195
waypoint7:2.65316
waypoint8:3.10207
waypoint9:3.60476
}\baselineposedropoutu

\pgfplotstableread[col sep=colon]{
wp:err
waypoint0:0.390215
waypoint1:0.690227
waypoint2:0.990229
waypoint3:1.34827
waypoint4:1.73195
waypoint5:2.14574
waypoint6:2.58641
waypoint7:3.06585
waypoint8:3.58312
waypoint9:4.14265
}\fullu

\pgfplotstableread[col sep=colon]{
wp:err
waypoint0:0.369605
waypoint1:0.97979
waypoint2:1.54291
waypoint3:1.97861
waypoint4:2.29052
waypoint5:2.53807
waypoint6:2.82508
waypoint7:3.17095
waypoint8:3.56696
waypoint9:4.0046
}\baselineposedropoutp

\pgfplotstableread[col sep=colon]{
wp:err
waypoint0:0.284389
waypoint1:0.67208
waypoint2:1.06587
waypoint3:1.37682
waypoint4:1.62336
waypoint5:1.87539
waypoint6:2.16403
waypoint7:2.50848
waypoint8:2.91514
waypoint9:3.36523
}\baselineposedropoutperturbp

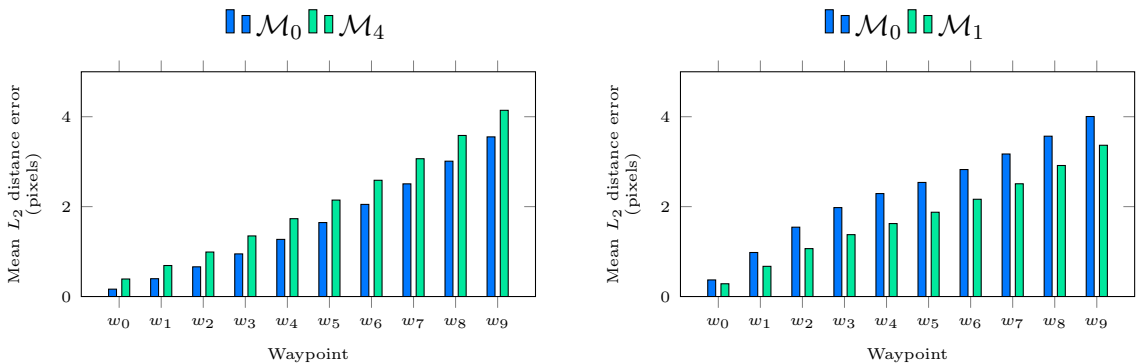
\begin{figure*}[t]
\centering
\subcaptionbox{Prediction Error for models $\M{0}$ and $\M{4}$ on unperturbed evaluation data.\label{fig:mse}}{
\begin{tikzpicture}
    \begin{axis}[
            ybar,
            width=0.5\textwidth,
            height=.3\textwidth,
            xlabel=Waypoint,
            ylabel style={align=center},
            ylabel=Mean $L_2$ distance error\\(pixels),
            symbolic x coords={waypoint0,waypoint1,waypoint2,waypoint3,waypoint4,waypoint5,waypoint6,waypoint7,waypoint8,waypoint9},
            xticklabels={$w_0$,$w_1$,$w_2$,$w_3$,$w_4$,$w_5$,$w_6$,$w_7$,$w_8$,$w_9$},
            xtick=data,
            xticklabel style={font=\tiny},
            yticklabel style={font=\tiny},
            xlabel style={font=\tiny},
            ylabel style={font=\tiny},
            ymin=0, ymax=5,
            bar width=3pt,
            legend style={at={(0.5,1.2)},anchor=center,legend columns=2,draw=none},
        ]
        \addplot [fill=wblue] table[x=wp,y=err]{\baselineu};
        \addplot [fill=wgreen] table[x=wp,y=err]{\fullu};
        \legend{$\M{0}$,$\M{4}$}
    \end{axis}
\end{tikzpicture}}
\hfill
\subcaptionbox{Prediction Error for models $\M{0}$ and $\M{1}$ on perturbed evaluation data.\label{fig:perturbed_mse}}{
\begin{tikzpicture}
    \begin{axis}[
            ybar,
            width=0.5\textwidth,
            height=.3\textwidth,
            xlabel=Waypoint,
            ylabel style={align=center},
            ylabel=Mean $L_2$ distance error\\(pixels),
            symbolic x coords={waypoint0,waypoint1,waypoint2,waypoint3,waypoint4,waypoint5,waypoint6,waypoint7,waypoint8,waypoint9},
            xticklabels={$w_0$,$w_1$,$w_2$,$w_3$,$w_4$,$w_5$,$w_6$,$w_7$,$w_8$,$w_9$},
            xtick=data,
            xticklabel style={font=\tiny},
            yticklabel style={font=\tiny},
            xlabel style={font=\tiny},
            ylabel style={font=\tiny},
            ymin=0, ymax=5,
            bar width=3pt,
            legend style={at={(0.5,1.2)},anchor=center,legend columns=2,draw=none},
        ]
        \addplot [fill=wblue] table[x=wp,y=err]{\baselineposedropoutp};
        \addplot [fill=wgreen] table[x=wp,y=err]{\baselineposedropoutperturbp};
        \legend{$\M{0}$,$\M{1}$}
    \end{axis}
\end{tikzpicture}}
\caption{Open loop evaluation results.}
\end{figure*}

\subsection{Open Loop Evaluation}
In an open-loop evaluation, we take test examples of expert driving data
and for each example, compute the $L_2$ distance error between the predicted and ground-truth waypoints. Unlike the
closed-loop setting, the predictions are not used to drive the agent forward and thus the network
never sees its own predictions as input. \cref{fig:mse} shows the $L_2$ distance metric in this open-loop
evaluation setting for models $\M{0}$ and $\M{4}$ on a test set of 10,000 examples.
These results show that model $\M{0}$ makes fewer errors than the full model $\M{4}$, but we know from
closed-loop testing that $\M{4}$ is a far better driver than $\M{0}$.
This shows how open-loop evaluations can be misleading, and closed-loop evaluations are critical while
assessing the real performance of such driving models.

\begin{figure*}[t]
\centering
\subcaptionbox{Ground-truth\label{fig:perturb_gt}}[0.32\textwidth]{
  \includegraphics[width=0.32\textwidth]{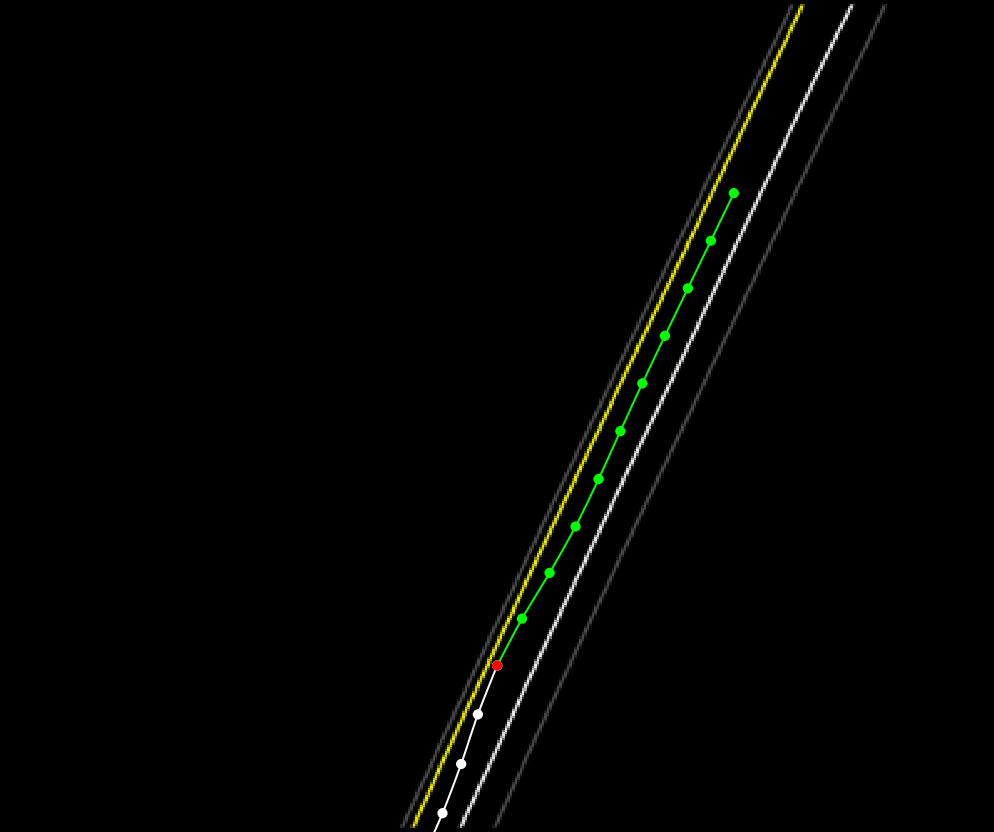}
  \includegraphics[width=0.32\textwidth]{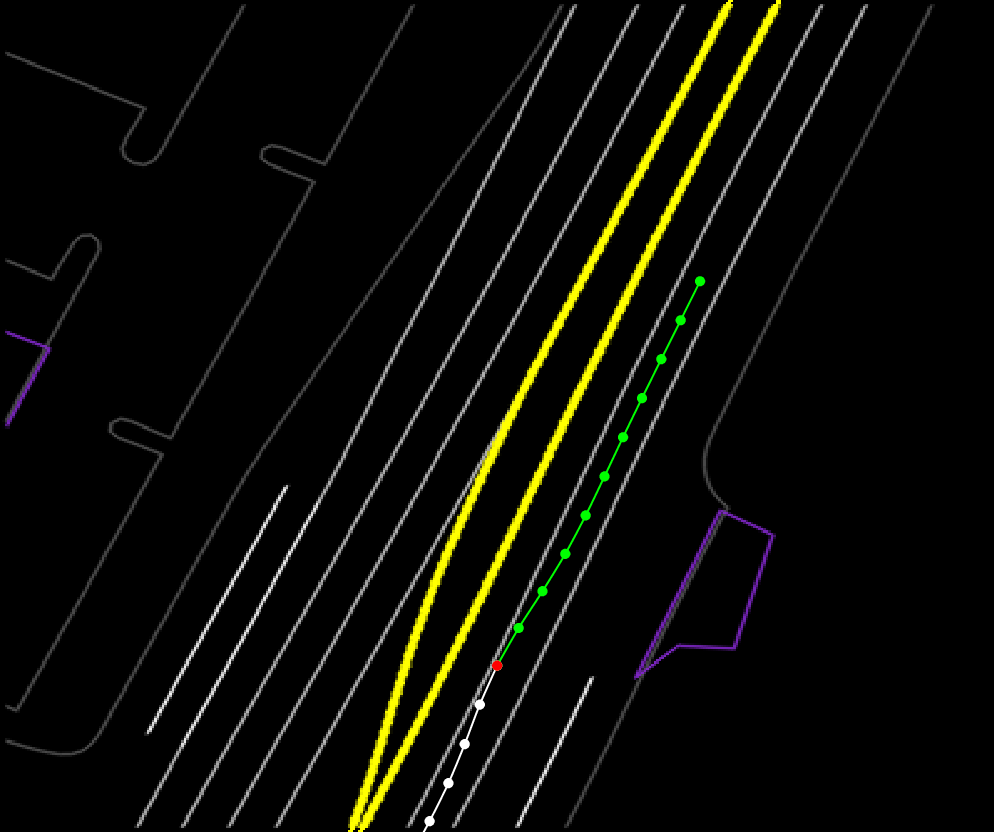}
}
\subcaptionbox{Model $\M{0}$ Prediction\label{fig:perturb_m0}}[0.32\textwidth]{
  \includegraphics[width=0.32\textwidth]{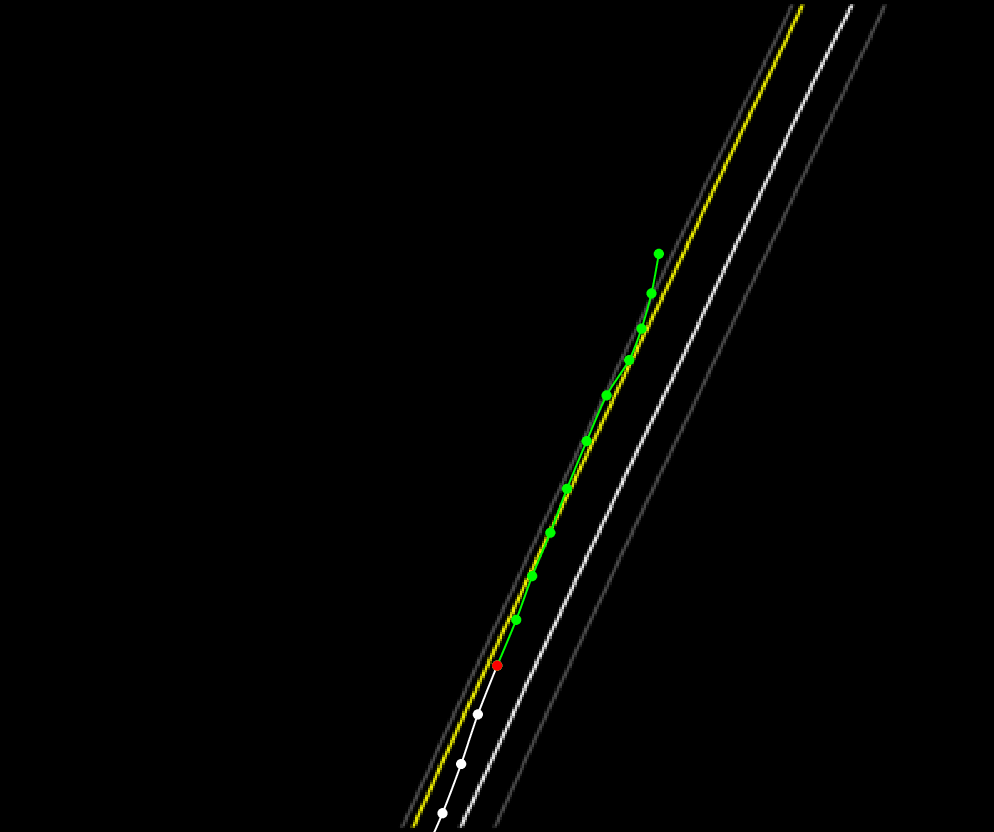}
  \includegraphics[width=0.32\textwidth]{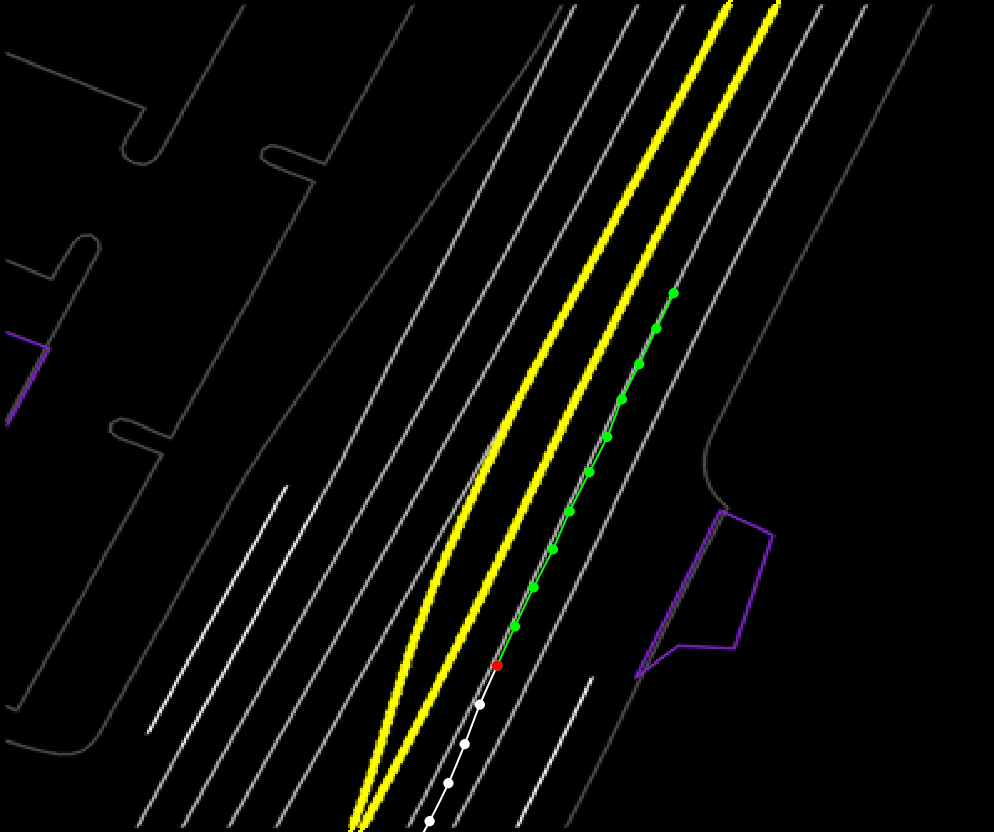}
}
\subcaptionbox{Model $\M{1}$ Prediction\label{fig:perturb_m1}}[0.32\textwidth]{
  \includegraphics[width=0.32\textwidth]{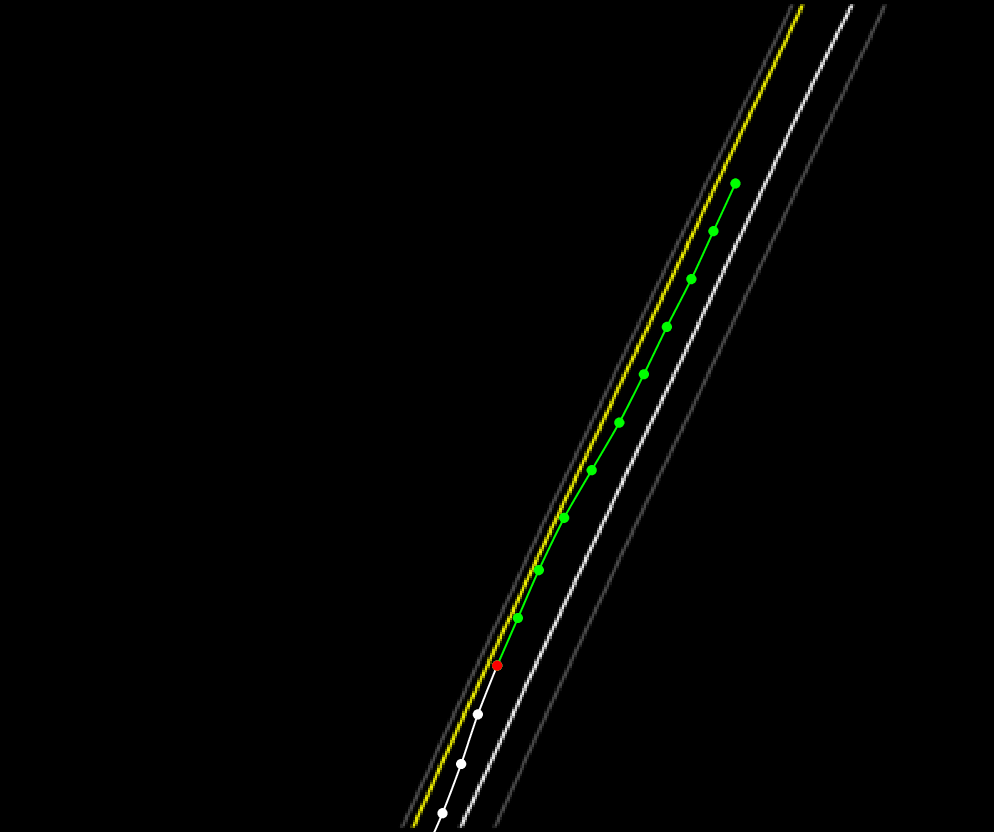}
  \includegraphics[width=0.32\textwidth]{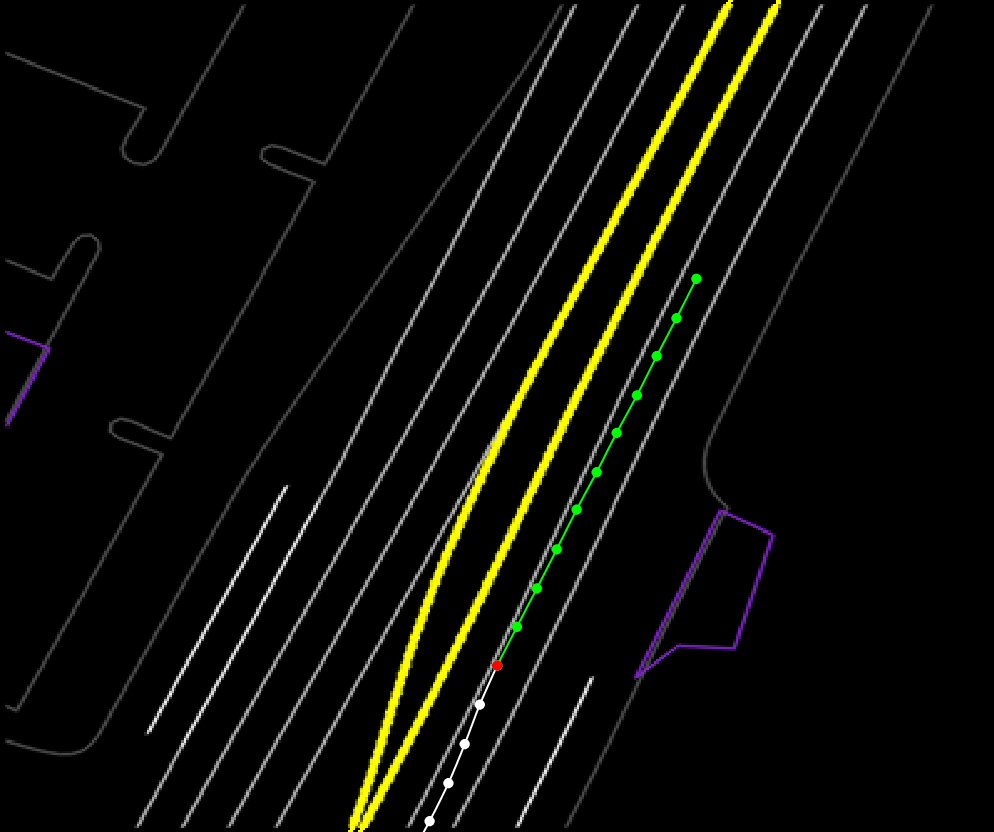}
}
\caption{Comparison of ground-truth trajectory in (a) with the predicted trajectories from models $\M{0}$ and $\M{1}$ in (b) and (c) respectively
on two perturbed examples. The red point is the reference pose $(u_0, v_0)$, white points are the past poses and green points are the future poses.\label{fig:perturbed_pred}}
\end{figure*}

We also compare the performance of models $\M{0}$ and $\M{1}$ on our perturbed evaluation data w.r.t
the $L_2$ distance metric in \cref{fig:perturbed_mse}. Note that the model trained without including
perturbed data ($\M{0}$) has larger errors due to its inability to bring the agent back from the
perturbation onto its original trajectory. \cref{fig:perturbed_pred} shows examples of the trajectories predicted
by these models on a few representative examples showcasing that the perturbed data is critical to avoiding
the veering-off tendency of the model trained without such data.

\subsection{Failure Modes}
At our ground resolution of $20$ cm/pixel, the agent currently sees $64$ m in front and $40$ m on the
sides and this limits the model's ability to perform merges on T-junctions and turns from a high-speed road.
Specific situations like U-turns and cul-de-sacs are also not currently handled, and will require sampling enough
training data. The model occasionally gets stuck in some low speed nudging situations. It sometimes outputs turn geometries
that make the specific turn infeasible (e.g. large turning radius). We also see some cases where the model gets over
aggressive in novel and rare situations for example by trying to pass a slow moving vehicle. We believe that adequate simulated exploration may be
needed for highly interactive or rare situations. 

\subsection{Sampling Speed Profiles}
\begin{figure*}[!t]
\centering
\begin{centering}
\subcaptionbox{$\log{P_1(x,y)}$}{\includegraphics[width=0.2\textwidth]{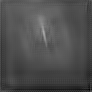}}
\subcaptionbox{$\log{P_5(x,y)}$}{\includegraphics[width=0.2\textwidth]{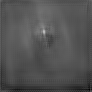}}
\end{centering}
\caption{Sampling speed profiles. The probability distribution $P_1(x,y)$ predicted
by the model at timestep $k=1$ allows us to sample different speed profiles conditioned
on which the later distribution $P_5(x,y)$ gets more constrained.}
\label{fig:speed}
\end{figure*}

The waypoint prediction from the model at timestep $k$ is represented by the
probability distribution $P_k(x,y)$ over the spatial domain in the top-down
coordinate system. In this paper, we pick the
mode of this distribution $\vec{p}_k$ to update the memory of the $AgentRNN$.
More generally, we can also sample from this distribution to allow us to predict
trajectories with different speed profiles. \cref{fig:speed}
illustrates the predictions $P_1(x,y)$ and $P_5(x,y)$ at the first and the fifth
iterations respectively, for a training example where the past motion history has
been dropped out. Correspondingly, $P_1(x,y)$ has a high uncertainity along the
longitudinal position and allows us to pick from a range of speed samples. Once
we pick a specific sample, the ensuing waypoints get constrained
in their ability to pick different speeds and this shows as a centered distribution
at the $P_5(x,y)$.

The use of a probability distribution over the next waypoint also presents
the interesting possibility of constraining the model predictions at inference time
to respect hard constraints. For example, such constrained sampling may provide
a way to ensure that any trajectories we generate strictly obey legal restrictions
such as speed limits. One could also constrain sampling of trajectories to a
designated region, such as a region around a given reference trajectory.

\section{Discussion}
In this paper, we presented our experience with what it took to get imitation learning to perform well in real-world driving. We found that key to its success is synthesizing interesting situations around the expert's behavior and augmenting
appropriate losses that discourage undesirable behavior. This constrained
exploration is what allowed us to avoid collisions and off-road driving even though
such examples were not explicitly present in the expert's demonstrations. To support it, and to best leverage the expert data, we used middle-level input and output
representations which allow easy mixing of real and simulated data and alleviate the burdens of learning perception and control.
With these ingredients, we got a model good enough to drive a real car.
That said, the model is not yet fully competitive with motion planning approaches but we feel that this is a good
step forward for machine learned driving models. There is room for improvement: comparing to end-to-end approaches, and
investigating alternatives to imitation dropout are among them. But most importantly, we believe that augmenting the expert demonstrations with
a thorough exploration of rare and difficult scenarios in simulation, perhaps within a
reinforcement learning framework, will be the key to improving the performance of these models especially for highly interactive scenarios.

\acks{We would like to thank Andrew Barton-Sweeney for help running the model on the car, Aleksandar Gabrovski for help with the simulation, Anca Dragan and Dragomir Anguelov for reviewing the paper and suggesting several improvements.
}


\bibliography{end2end}

\end{document}